\pdfoutput=1
\documentclass{article}
\usepackage{spconf,amsmath,graphicx} 
\usepackage{hyperref}
\usepackage{cite}
\usepackage{amsmath,amssymb,amsfonts}
\usepackage{algorithmic}
\usepackage{graphicx}
\usepackage{textcomp}
\usepackage{xcolor}
\usepackage{graphicx}
\usepackage{svg}
\usepackage{multirow}                 
\usepackage{multicol}                 
\usepackage{multirow}                
\usepackage{float}                    
\usepackage{makecell}                 
\usepackage{booktabs}                 
\usepackage{amsmath}
\usepackage{amssymb}
\usepackage{enumitem}
\setlist{nosep, leftmargin=14pt}

\usepackage{mwe} 

\newcommand{\capitalize}[1]{%
    \begingroup
    \edef\@{\lowercase{#1}}%
    \uppercase{\expandafter\def\expandafter\@\expandafter{\@}}%
    \expandafter\endgroup\@
}
\title{Confounder-Aware Medical Data Selection for Fine-Tuning Pretrained Vision Models
\thanks{C\capitalize{ode is available at: \href{https://github.com/MAXXXN/Confounder-Aware-Medical-Data-Selection}{https://github.com/MAXXXN/Confounder-Aware-Medical-Data-Selection} }}}
\name{
\parbox{\textwidth}{
\centering
Anyang Ji$^{1,3}$ 
\qquad Qingbo Kang$^{1}$ 
\qquad Wei Xu$^{1,4}$ \\
Changfan Wang$^{2}$ 
\qquad Kang Li$^{*1}$ 
\qquad Qicheng Lao$^{*2}$
\thanks{* Corresponding authors: likang@wchscu.cn; qicheng.lao@bupt.edu.cn}
}
}
\address{$^{1}$ West China Biomedical Big Data Center, West China Hospital, Sichuan University, Chengdu, China \\
    $^{2}$ School of Artificial Intelligence, Beijing University of Posts and Telecommunications, Beijing, China \\
    $^{3}$ College of Computer Science, Sichuan University, Chengdu, China \\
    $^{4}$ School of Biomedical Engineering, University of Science and Technology of China, Hefei, China}
\begin{document}
%
\maketitle
\begin{abstract}
The emergence of large-scale pre-trained vision foundation models has greatly advanced the medical imaging field through the pre-training and fine-tuning paradigm. However, selecting appropriate medical data for downstream fine-tuning remains a significant challenge considering its annotation cost, privacy concerns, and the detrimental effects of confounding variables. 
In this work, we present a confounder-aware medical data selection approach for medical dataset curation aiming to select minimal representative data by strategically mitigating the undesirable impact of confounding variables while preserving the natural distribution of the dataset. 
Our approach first identifies confounding variables within data and then develops a distance-based data selection strategy for confounder-aware sampling with a constrained budget in the data size. 
We validate the superiority of our approach through extensive experiments across diverse medical imaging modalities, highlighting its effectiveness in addressing the substantial impact of confounding variables and enhancing the fine-tuning efficiency in the medical imaging domain, compared to other data selection approaches. 
\end{abstract}
\begin{keywords}
data selection, confounding variables, fine-tuning
\end{keywords}
\section{Introduction}
\label{sec:intro}
The emergence of large-scale pre-trained vision foundation models~\cite{dosovitskiy2020image,caron2021emerging,radford2021learning,he2022masked} 
has significantly enhanced the downstream task performance, particularly when integrated with a fair amount of data for fine-tuning. However, acquiring, annotating, and managing these fine-tuning data in the medical domain still entail substantial costs and demand increased computational resources. In medical imaging, these issues are further exacerbated by the stringent confidentiality requirements of healthcare data. Amassing extensive medical imaging datasets places considerable strain on resources and heightens the risk of data breaches, raising serious concerns for patient confidentiality~\cite{LV2020103}. Therefore, selecting minimal representative data for fine-tuning represents a significant research direction~\cite{toneva2018empirical,paul2021deep,xia2022moderate}.

Data selection aims to identify the most informative data within a large dataset to constitute a subset, commonly known as a \textit{coreset}. An effective data selection strategy is vital not only for reducing the associated risks of utilizing large datasets in sensitive areas but also for enhancing the performance and versatility of models trained on heterogeneous datasets. Recently, research on data selection for deep learning tasks has emerged, grounded in various definitions of sample significance. Toneva $et\ al.$~\cite{toneva2018empirical} propose a forgetting score 
and remove samples that are less prone to being forgotten.  Paul $et\ al.$~\cite{paul2021deep} develop two scores: the Gradient Normed (GraNd) and the Error L2-Norm (EL2N), which evaluate the norm of the gradient and the L2-distance between the normalized error from predicted probabilities and the corresponding one-hot labels, to isolate noteworthy samples. 
Xia $et\ al.$~\cite{xia2022moderate} select samples with representations near the median distance from the class center.
However, although numerous works have been proposed for data selection, only a few methods cater specifically to medical imaging~\cite{kim2020confident,he2023data}, and the majority fail to incorporate its distinctive characteristics, such as susceptible to confounding variables~\cite{vanderweele2013definition,pourhoseingholi2012control}.

\begin{figure*}[!t]
\centering
\includegraphics[width=0.92\textwidth]{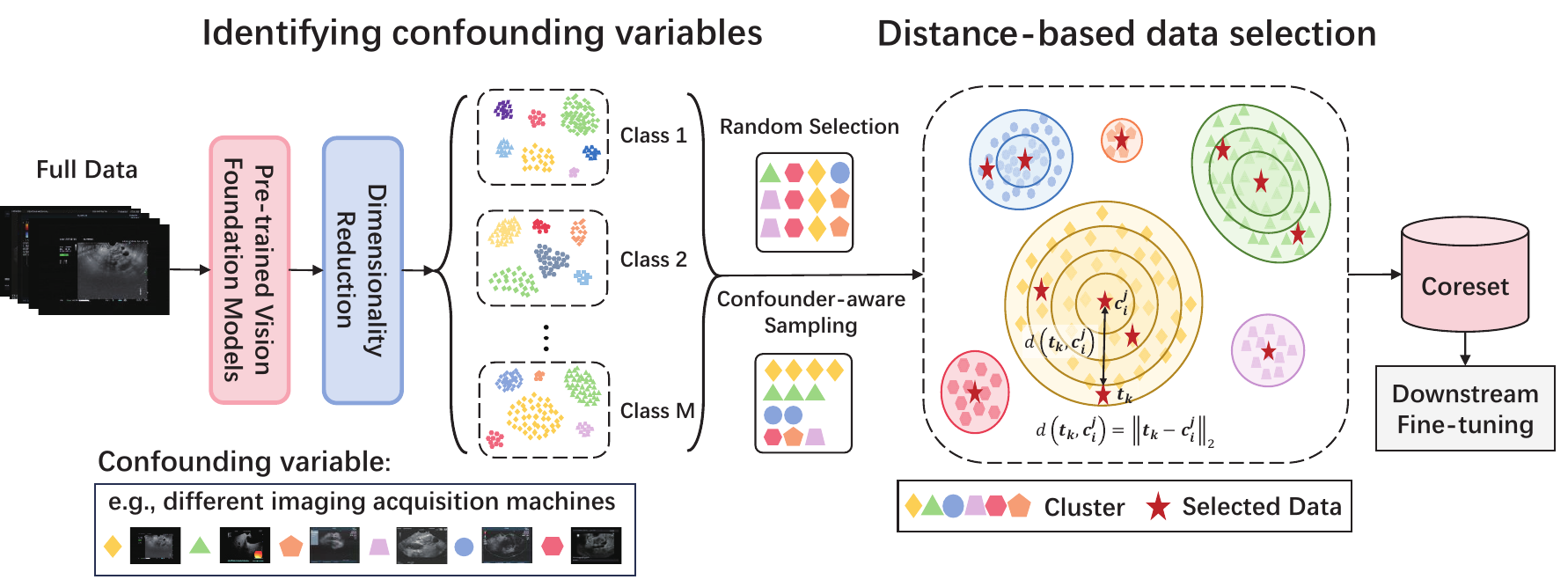}
\caption{The framework of our proposed method based on the confounding variable.} 
\label{fig1}
\end{figure*}
In medical imaging, the complex nature of medical imaging datasets may easily introduce confounding variables. 
These confounders
can introduce systematic biases into the dataset, leading to misleading correlations and erroneous conclusions~\cite{jager2008confounding,he2019practical}. 
As confounders not only reduce the performance
but also pose the risk of compromising patient care by propagating incorrect medical assessments, addressing them is crucial for effective and efficient data selection in the medical domain. 

To address the aforementioned issues, in this paper, we propose a confounder-aware medical data selection approach based on confounder-aware sampling, aiming to mitigate the effects of confounders. Specifically, our approach comprises two main stages. In the first stage, confounding variables that may introduce potential biases to task performance, are identified using pre-trained vision foundation models and feature clustering. 
In the second stage, a distance-based data selection method for confounder-aware sampling is introduced based on the identified confounding variables, 
where samples are selected based on their relative distance from the centers of the confounder clusters in a uniform manner throughout the entire pipeline. 
Our findings demonstrate that 
by strategically controlling for confounders, a carefully curated dataset may surpass the performance of larger but less discerning datasets. This highlights the significant, though often overlooked, impact of confounders in data selection. 
In summary, our contributions are as follows:
\begin{itemize}
    \item We propose a data selection approach that identifies confounders and employs a distance-based strategy to preserve the data's natural distribution while minimizing confounder bias.
    \item We establish the connection between confounders and medical imaging data selection, demonstrating that curated datasets can outperform larger ones when confounders are controlled.
    \item We conduct experiments on datasets with three different imaging modalities for downstream classification fine-tuning tasks, and the results demonstrate that our method achieves the state-of-the-arts performance.
\end{itemize}

\begin{figure*}[!t]
\centering
\includegraphics[width=0.9\textwidth]{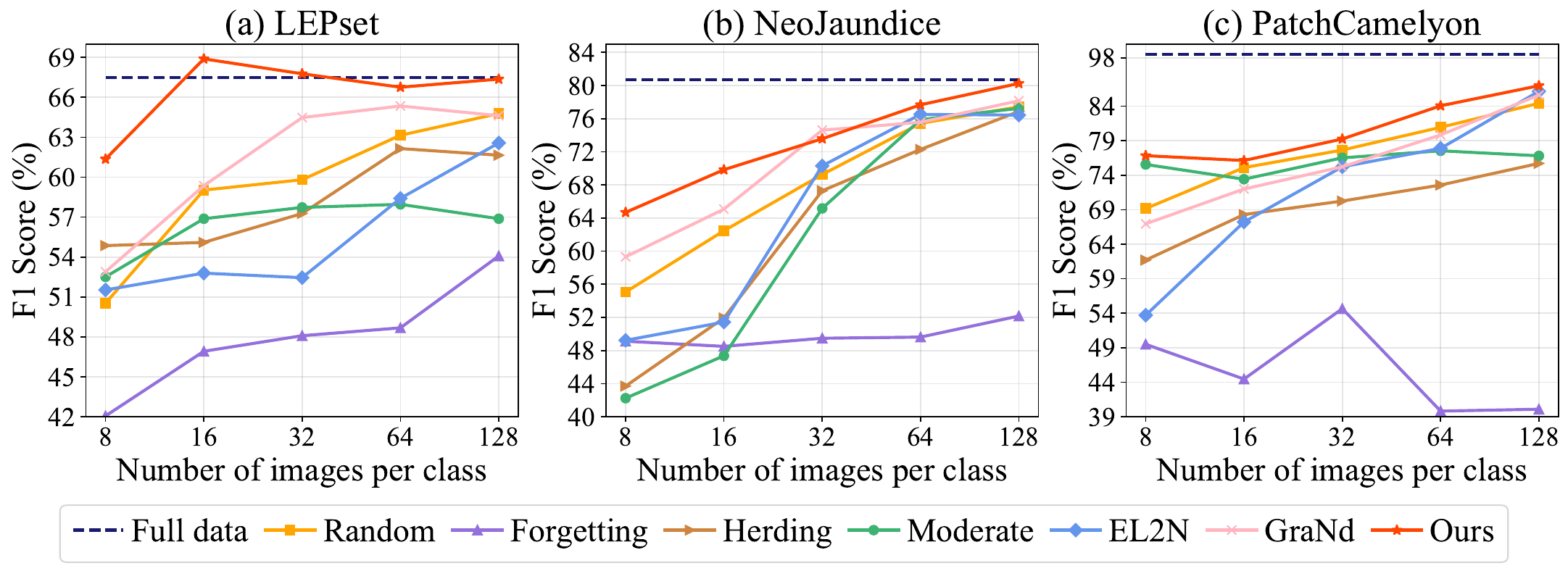}
\caption{Comparisons of our proposed method with various data selection approaches on (a) LEPset~\cite{li_2023_8041285}, (b) NeoJaundice~\cite{Wang2023}, and (c) PatchCamelyon~\cite{b_s_veeling_j_linmans_j_winkens_t_cohen_2019_2546921}.} \label{fig2}
\end{figure*}

\section{Method}

Confounding variables have the potential to obscure underlying pathological processes, resulting in a model with diminished generalizability and increased susceptibility to errors when applied to novel or external datasets~\cite{jager2008confounding}. Therefore, controlling for confounders is crucial. In this section, we elaborate on our proposed confounder-aware data selection approach. The overall framework is depicted in Fig.~\ref{fig1}, which consists of two principle stages: identifying confounding variables and performing distance-based data selection based on the identified confounders.

\subsection{Identifying Confounding Variables} 
As depicted in Fig.~\ref{fig1}, the identification of confounding variables involves three main steps: feature extraction based on pre-trained vision models, dimensionality reduction, and class-wise clustering.  

\textbf{Feature Extraction:} 
Initially, to distill essential data attributes for more effective identification of confounding variables, we utilize pre-trained vision foundation models for feature extraction. Given the full training data $\mathcal{D} = \{{\mathbf{s}_1, \mathbf{s}_2,\ldots, \mathbf{s}_n}\}$, the pre-trained encoder transforms all the data points into the latent representations \( \{{\mathbf{z}_1, \mathbf{z}_2,\ldots, \mathbf{z}_n}\} \in \mathcal{Z} \) by executing a sequence of transformations. These transformations are typically implemented as a neural network with parameters \( \theta \), which remain fixed during this operation:
\begin{equation}
\mathbf{z}_i = f_{\text{enc}}(\mathbf{x}_i; \theta) ,
\end{equation}
In our method, we adopt Masked Autoencoder (MAE)~\cite{he2022masked} for feature extraction. Specifically, the MAE encoder pre-trained on the ImageNet~\cite{deng2009imagenet} is utilized to transform input images into informative latent representations. 

\textbf{Dimensionality Reduction and Clustering:} 
To facilitate the visualization of high-dimensional features and thereby identify the confounding variables, our approach incorporates dimensionality reduction techniques.
We represent the dimensionality reduction process as follows:
\begin{equation}
\mathbf{t}_i = f_{\text{d}}(\mathbf{z}_i),
\end{equation}
where $\mathbf{t}_i\in\mathbb{R}^{2}$ denotes the representation of $\mathbf{z}_i$ after dimensionality reduction.


As shown in Fig.\ref{fig1}, after dimensionality reduction, a distance metric is computed to partition data points into clusters based on confounding variables. Using an endoscopic ultrasonography dataset (LEPset\cite{li_2023_8041285}), we illustrate clustering by ultrasound machine variance as the confounder. DBSCAN~\cite{ester1996density} is employed to support the sampling process.

\subsection{Distance-based Data Selection}
As shown in Fig.~\ref{fig1}, random sampling can introduce bias by over-representing certain clusters and neglecting others, distorting the data distribution. To mitigate this, we propose a distance-based data selection method with confounder-aware sampling, preserving the natural distribution while reducing confounder influence.

Specifically, after identifying confounding variables,  the center of the $i$-th cluster belonging to class $j$ can be defined as:
\begin{equation}
\mathbf{c}^{j}_i = \frac{1}{|\mathbf{T}^{j}_i|} \sum_{\mathbf{t}_k \in \mathbf{T}^{j}_i} \mathbf{t}_k, 
\end{equation}
where $\mathbf{T}^{j}_i$ denotes all data points assigned to this cluster, and $|\mathbf{T}^{j}_i|$ signifies the cardinality of the vector set $\mathbf{T}^{j}_i$. We can easily compute the Euclidean distance between the data point $\mathbf{t}_k \in \mathbf{T}^{j}_i$ and the corresponding cluster center as follows:
\begin{equation}
d(\mathbf{t}_k, \mathbf{c}^{j}_i) = \left \| \mathbf{t}_k - \mathbf{c}^{j}_i \right \|_2,
\end{equation}

Subsequently, the points within each cluster are arranged in ascending order based on their distance from the center, yielding an ordered sequence denoted as $\mathbf{S}^{j}_i$. Let $n^{j}_i$ represent the number of data points uniformly sampled from $\mathbf{T}^{j}_i$, which is determined proportionally to the size of $\mathbf{T}^{j}_i$:
\begin{equation}
n^{j}_i = \left\lfloor \frac{|\mathbf{T}^{j}_i|}{\sum_{k=1}^{M} |\mathbf{T}^{j}_k|} \cdot N \right\rfloor + b ,
\end{equation}
where $\lfloor \cdot \rfloor$ means round down, and $N$ is the budget number of sampled data for each class. To ensure accuracy, we validate the computed $n^{j}_i$. If the number of $\sum n^{j}$ is less than N, we adjust the sampling number by setting $b = N-\sum n^{j}$ for the cluster with the greatest size.

Finally, the set of sampled data points, denoted as $\mathbf{U}^{j}_i$, can be represented as follows:
\begin{equation}
\mathbf{U}^{j}_i(n^{j}_i) = \left\{\mathbf{S}^{j}_i\left[\left\lfloor \frac{k \cdot |\mathbf{S}^{j}_i|}{n^{j}_i} \right\rfloor\right] , k = 0, 1, \ldots, n^{j}_i-1 \right\} .
\end{equation}
where $\mathbf{U}$ is considered the selected coreset $\mathcal{D_S}\subset\mathcal{D}$ for downstream fine-tuning.

\section{Experiments}

\subsection{Experimental Setup}

\subsubsection{Datasets and Architectures} 
We evaluate our method on three datasets with different imaging modalities: LEPset~\cite{li_2023_8041285} (3,500 endoscopic ultrasonography (EUS) images for pancreatic cancer classification), NeoJaundice~\cite{Wang2023} (2,235 digital camera images for neonatal jaundice classification), and PatchCamelyon~\cite{b_s_veeling_j_linmans_j_winkens_t_cohen_2019_2546921} (220,025 pathological images for identifying metastatic tissues in lymph nodes). Each dataset is randomly split into training, validation, and test sets at a 7:1.5:1.5 ratio, with coresets containing 8, 16, 32, 64, and 128 samples per class.
For downstream fine-tuning, we employ a ViT-Base~\cite{dosovitskiy2020image} architecture pre-trained with MAE on ImageNet.

\subsubsection{Implementation Details} 
We implement our approach with PyTorch. The image size for both data selection and downstream fine-tuning is 224 $\times$ 224. We set the batch size to 8 for all experiments. An AdamW optimizer with a weight decay factor of 0.05 and a base learning rate of 1e-4 is used. F1 score is utilized as the classification performance evaluation metric. All models were trained for 500 epochs and the model with the highest performance on the validation set is selected for evaluation.

\subsection{Results}
\subsubsection{Comparison with the state-of-the-arts} We compare our approach with several commonly used data selection methods, including Random Sampling, Forgetting~\cite{toneva2018empirical}, Herding~\cite{welling2009herding}, Moderate~\cite{xia2022moderate}, EL2N~\cite{paul2021deep}, and GraNd~\cite{paul2021deep}. 

As shown in Fig.~\ref{fig2}, our method consistently outperforms all competing methods across different numbers of images per class on all three datasets. Furthermore, at smaller coreset sizes (8 and 16 images per class), our approach demonstrates a significant performance boost, underscoring its effectiveness in low-data scenarios common in medical imaging.


Notably, for the LEPset dataset, our method's performance with 16 images per class even surpasses that achieved with the full training data.
This finding indicates that quantity does not always equate to quality, 
highlighting the often-overlooked impact of confounders on medical imaging. 

\subsubsection{Ablation Study} We conduct ablation studies encompassing three aspects: the significance of identifying confounding variables, data selection strategies, and pre-trained vision foundation models for feature extraction. 

\begin{figure}[t]
\centering
\includegraphics[width=0.44\textwidth]{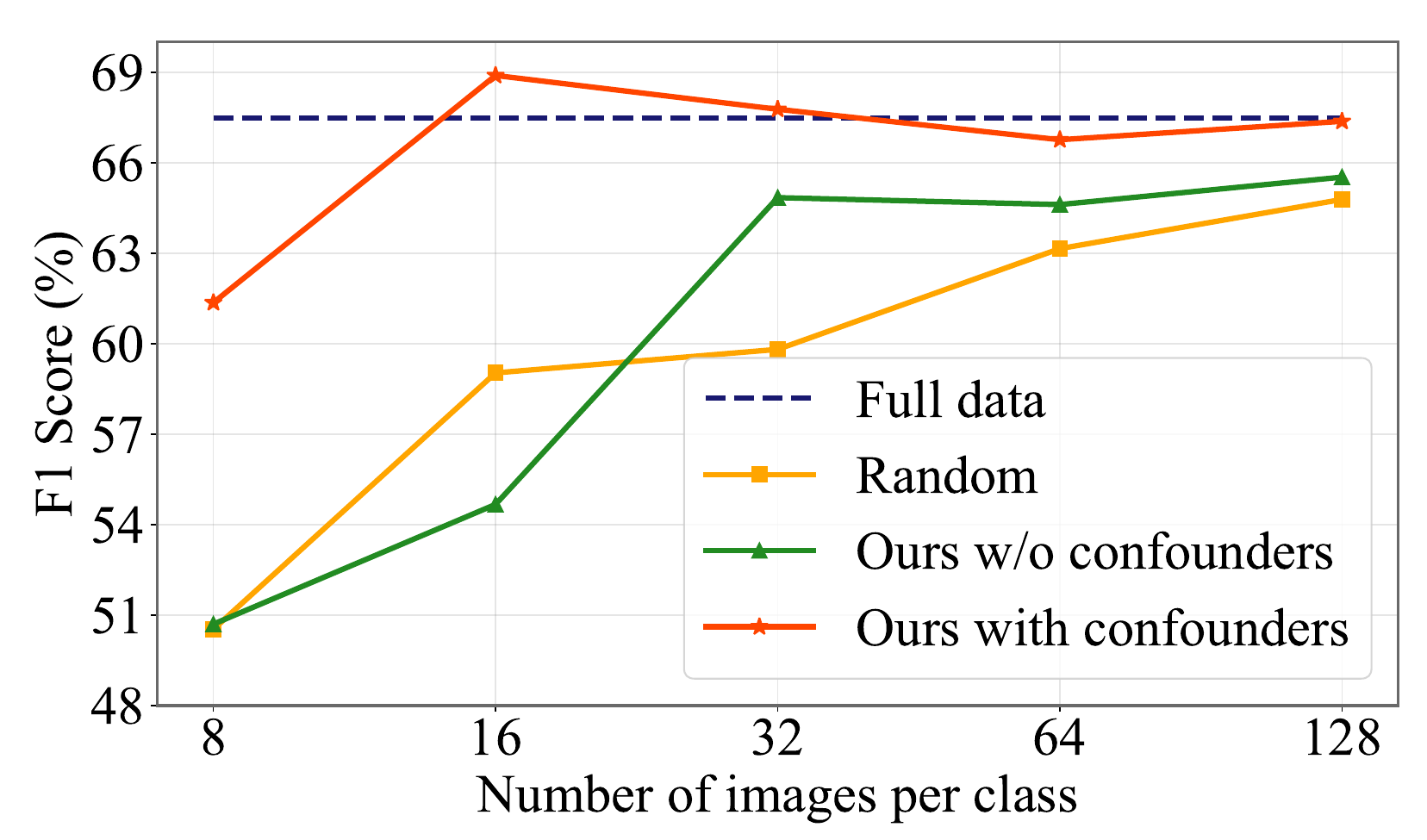}
\caption{Ablation study on the significance of identifying confounding variables.}\label{fig5}
\end{figure}

\textbf{The Significance of Identifying Confounding Variables:} 
To evaluate the significance of confounding variables, we compare our method with and without confounders. As depicted in Fig.~\ref{fig5}, our confounder-aware method outperforms other approaches, including the one without confounders, demonstrating that confounding variable identification can bring significant performance improvement.

\textbf{Data Selection Strategies:} 
We experiment with four data selection strategies: (1) selecting points closest to the cluster center, (2) furthest from the center, (3) at a medium distance, and (4) uniform sampling, which is our final choice. Table~\ref{tab2} shows that uniform sampling delivers the best performance.

\begin{table}[!t]
\renewcommand\arraystretch{0.85}
\begin{minipage}[t]{0.48\textwidth}
\makeatletter\def\@captype{table}\makeatother\caption{Ablation study on data selection strategies (\%).}\label{tab2}
\centering
\setlength{\tabcolsep}{4pt}
\fontsize{9}{12}\selectfont
\begin{tabular}{ccccccc}
\toprule
\multirow{2}*{Images per class} & \multicolumn{2}{c}{8} & \multicolumn{2}{c}{32} & \multicolumn{2}{c}{128}\\
\cmidrule(lr){2-3}\cmidrule(lr){4-5}\cmidrule(lr){6-7}
  & Acc & F1 & Acc & F1 & Acc & F1\\

\midrule
Random & 55.20 & 50.53 & 61.33 & 59.81 & 63.34 & 64.78\\
\midrule
Closest & 56.05 & 55.77 & 65.86 & 65.76 & 66.23 & 66.18\\
Furthest & 51.06 & 45.38 & 60.77 & 60.47 & 65.69 & 65.52\\
Median & \pmb{61.74} & 59.83 & 66.43 & 66.36 & 66.17 & 66.10\\
\pmb{Uniform} & 61.57 & \pmb{61.36} & \pmb{67.94} & \pmb{67.77} & \pmb{67.66} & \pmb{67.37}\\
\bottomrule
\end{tabular}
\end{minipage}
\hspace{1mm} 
\begin{minipage}[t]{0.48\textwidth}
\makeatletter\def\@captype{table}\makeatother\caption{Ablation study on feature extraction models (\%).}\label{tab3}
\centering
\setlength{\tabcolsep}{4pt}
\fontsize{9}{12}\selectfont
\begin{tabular}{ccccccc}
\toprule
\multirow{2}*{Images per class} & \multicolumn{2}{c}{8} & \multicolumn{2}{c}{32} & \multicolumn{2}{c}{128}\\
\cmidrule(lr){2-3}\cmidrule(lr){4-5}\cmidrule(lr){6-7}
  & Acc & F1 & Acc & F1 & Acc & F1\\
\midrule
Random & 55.20 & 50.53 & 61.33 & 59.81 & 63.34 & 64.78\\
\midrule
ResNet~\cite{he2016deep} & 55.40 & 53.35 & 65.14 & 64.86 & 67.11 & 66.97\\
CLIP~\cite{radford2021learning} & 59.89 & 59.27 & 66.26 & 65.89 & 67.57 & 66.46\\
BiomedCLIP~\cite{zhang2024biomedclip} & 61.14 & 60.24 & 62.20 & 61.90 & 65.43 & 65.26\\
DINOv1~\cite{caron2021emerging} & 57.54 & 55.70 & 62.60 & 61.95 & 67.06 & 66.92 \\
\pmb{MAE}~\cite{he2022masked} & \pmb{61.57} & \pmb{61.36} & \pmb{67.94} & \pmb{67.77} & \pmb{67.66} & \pmb{67.37}\\
\bottomrule
\end{tabular}
\end{minipage}
\end{table}

\textbf{Feature Extraction Models:} 
Table~\ref{tab3} presents the results of using different pre-trained vision foundation models for feature extraction, including CLIP~\cite{radford2021learning}, BiomedCLIP~\cite{zhang2024biomedclip}, DINOv1~\cite{caron2021emerging}, and MAE. All models are pre-trained with ViT-Base~\cite{dosovitskiy2020image}.
MAE consistently achieves the best performance. While other models underperform compared to MAE, they still surpass the random sampling baseline, highlighting the robustness of our approach.

\section{Conclusion}
In this study, we propose a confounder-aware medical data selection approach for fine-tuning pre-trained foundation models. 
Our method utilizes pre-trained visual models and feature clustering to identify confounders and employs a distance-based strategy for data selection. This ensures the selected data subset is both representative and balanced, effectively diminishing the influence of confounding variables and preserving the dataset's inherent distribution. Comprehensive experiments validate the efficacy of our method. However, our approach has certain limitations, such as the requirement for manual calibration of clustering algorithm parameters to yield optimal alignment with anticipated clustering outcomes. 
We aim to address these limitations in future work. 

\bibliographystyle{IEEEbib}
\bibliography{strings,references}

\end{document}